\documentclass[acmsmall]{acmart}
\usepackage[english]{babel}
\usepackage{indentfirst}
\usepackage{multirow}
\usepackage{chngpage}
\usepackage{array}
\usepackage{graphicx}
\usepackage{color}
\usepackage{booktabs}
\usepackage{rotating}
\usepackage{times}
\usepackage{helvet}
\usepackage{courier}
\usepackage{url}
\usepackage{float}
\usepackage{graphicx}
\usepackage{epstopdf}
\usepackage{algorithm}
\usepackage{algorithmic}
\usepackage{amsmath}
\usepackage{amssymb}
\usepackage{multirow}
\usepackage{array}
\usepackage{subfigure}
\usepackage{colortbl,booktabs}
\usepackage{verbatim}
\AtBeginDocument{%
  \providecommand\BibTeX{{%
    \normalfont B\kern-0.5em{\scshape i\kern-0.25em b}\kern-0.8em\TeX}}}

\setcopyright{acmcopyright}
\copyrightyear{2020}
\acmYear{2020}

\acmJournal{TOMM}

\begin{document}

\title{Full Reference Screen Content Image Quality Assessment by Fusing Multi-level Structure Similarity}

\author{Chenglizhao Chen, Hongmeng Zhao, Huan Yang, Chong Peng* and Teng Yu}
\authornote{Chenglizhao Chen and Hongmeng Zhao contributed equally to this work.
Corresponding author: Chong Peng and Teng Yu.
Code\&Data is available from \url{https://github.com/HongmengZhao/SR-CNN}.}

\affiliation{%
  \institution{Qingdao University}}
\email{cclz123@163.com, zhm199411@163.com, cathy\_huanyang@hotmail.com, pchong1991@163.com, yutenghit@foxmail.com}


\author{Hong Qin}
\affiliation{%
  \institution{Stony Brook University}}
\email{qin@cs.stonybrook.edu}

\renewcommand{\shortauthors}{C. Chen and H. Zhao, et al.}

\begin{abstract}

The screen content images (SCIs) usually comprise various content
types with sharp edges, in which the artifacts or distortions can be well sensed by the vanilla structure similarity
measurement in a full reference manner.
Nonetheless, almost all of the
current SOTA structure similarity metrics are
``locally'' formulated in a single-level manner, while the true human
visual system (HVS) follows the multi-level manner, and such mismatch
could eventually prevent these metrics from achieving
trustworthy quality assessment.
To ameliorate, this paper advocates a novel solution to measure
structure similarity ``globally'' from the perspective of sparse
representation.
To perform multi-level quality assessment in
accordance with the real HVS, the above-mentioned global metric will
be integrated with the conventional local ones by resorting to the
newly devised selective deep fusion network. To validate its efficacy
and effectiveness, we have compared our method with 12
SOTA methods over two widely-used large-scale public SCI
datasets, and the quantitative results indicate that our method yields
significantly higher consistency with subjective quality score than
the currently leading works. Both the source code and data are also
publicly available to gain widespread acceptance and facilitate new
advancement and its validation.
\end{abstract}

\ccsdesc[500]{Computing methodologies~Multimedia}

\keywords{Image Quality Assessment; Screen Content Images; Selective Deep Fusion}
\maketitle

\section{Introduction and Motivation}

As one of the most informative representation venues, the screen
content images (SCIs) are ubiquitous in various multimedia devices,
such as mobile phone, laptop, tablet, and so
on~\cite{OurSPL18,Fang2017No,Miao2016A,OurPR16,OurPR15,Hu2014Toward}. From the
perspective of the multi-client communication systems, the SCIs received by multimedia may usually get distorted during acquisition, storage, transmission, and coding~\cite{Zhan2014Advanced,Zhu2015Hash,Chang2014Intra}
Thus, it is of very importance to design an effective quality assessment method
for SCIs.

In fact, the image quality assessment (IQA) problem has been paid
intensive research attention during the past several years. According
to the availability of the reference image, the state-of-the-art (SOTA) IQA
methods~\cite{Gu2017No,Zhou2018Local} can be roughly grouped into
three categories: the full reference, the reduced reference, and the
no reference methods~\cite{Qian2017Towards,Min2017Unified,2018FreitasBlind}, where the
first two of these three would need to know the full or partial
undistorted image respectively, while the last one would not. In order
to pay attention to this paper's main foci, we only shed light on the
topic of full reference with brief justification and discussion.

In essence, most of the existing full reference IQA methods are
designed for the natural scene images (NSIs). Nonetheless, different
from the NSIs which usually comprise image content with gradual
variations, the SCIs are frequently synthesized and rendered,
exhibiting image content with sharp
edges~\cite{Wang2016Objective,Wang2015Perceptual}. Furthermore, the
image content type of SCIs is also different from NSIs, i.e., the NSIs
mainly relate to pictures, while the SCIs contain both pictures and
textures.  Thus, the NSIs based methods are not feasible for the
SCIs~\cite{Ni2016Gradient}.

\begin{figure}[t]
\centering
\includegraphics[width=0.8\linewidth]{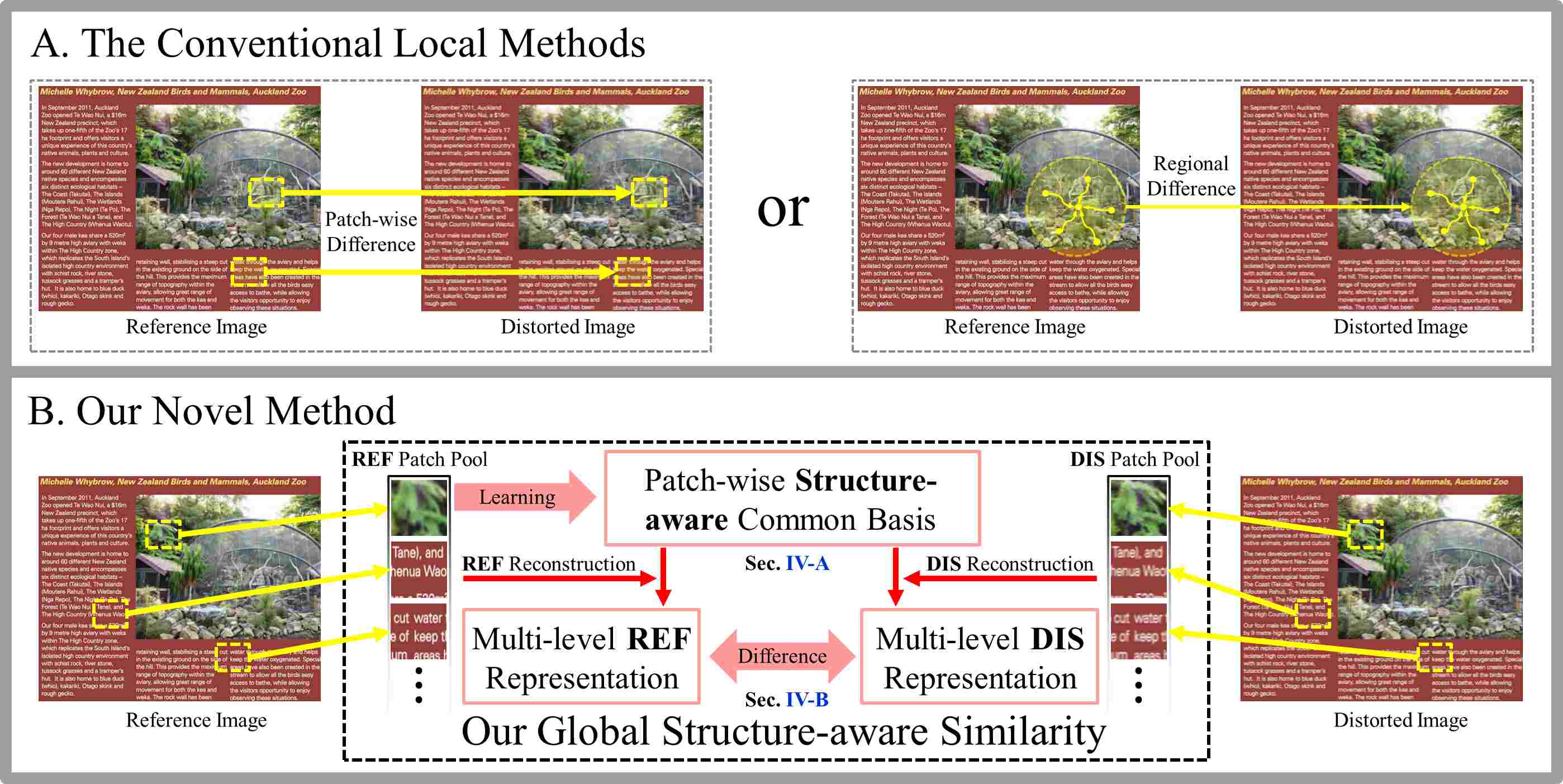}
\caption{The conventional methods frequently resort the single-level fashion for IQA, e.g., solely using patch-wise or regional-wise manner to sense distortion induced artifacts, which contradicts to the multi-level real HVS. In sharp contrast, this paper proposes to conduct measure the structure-aware similarity globally, which particulary suits the quality assessment for SCIs~\cite{Zhou2004Image,Gu2015Screen}. The abbreviation ``REF'' and ``DIS'' respectively denote the reference image and the distorted image.}
\label{fig:motivation}
\end{figure}

As for the current main stream full reference methods for SCIs, the
structure similarity~\cite{Zhou2004Image} has been widely adopted to
sense artifacts in distorted images, and a strong similarity between the reference SCI
and its distorted version usually indicates a high quality score, and
vice versa.
Although much more progresses have been made, almost all the structure
similarity measurement based variants are
locally formulated~\cite{Gu2016Saliency,Ni2017ESIM}, which could
easily produce untrustworthy quality assessment due to its limited
sensing scope (e.g., conducing IQA via patch-wise or regional-wise manner, please refer to
Fig.~\ref{fig:motivation}-A).
Moreover, most of the current SOTA works shall be classified into the single-level category, in which these works solely use either local or global measurement each time.
However, such ``single-level'' manner principally contradicts to the ``multi-level''
real human visual system (HVS) which shifts intermittently between the
entire image sweeping and the tiny artifact zooming, preventing the
SOTA methods from performing trustworthy quality
assessment.
Additionally, the image content type is another vital factor to
affect the overall performance~\cite{Zhang2018Quality}, which should
be considered when pursuing a trustworthy quality assessment.

Hence, all the above-mentioned arguments motivate us to investigate a
multi-level manner for the SCI quality assessment, which mainly
includes the following three aspects: low-level structure-aware
metrics, mid-level image content, and high-level selective fusion.
Meanwhile, since the SCIs can be attributed to a mixture of textures
and pictures, the HVS frequently pays more attention to the irregular
patterns (e.g., miss-aligned textures), thus the tiny structure
distortions may easily be observed by the HVS~\cite{ChenPR16,CC2019TMM1,CC2019TIP,CC2020TIP}. Therefore, it is
intuitive to choose the structure similarity as the ``low-level''
metric to detect distortion induced artifacts.
To compute the structure similarity globally, we utilize the sparse
representation~\cite{aharon2006k} to sequentially formulate the
structure consistency of the reference SCI as the global structure
dictionary.
By using this dictionary to respectively reconstruct the
reference SCI and its distorted version, we can use the variation degree toward dictionary usage to represent the global structure similarity (please refer to Fig.~\ref{fig:motivation}-B).
Meanwhile, we integrate the off-the-shelf local structure similarity into our newly
devised global one by using our newly designed ``high-level'' fusion
network.
To realize the ``mid-level'' content-aware attribute, we divide the target SCI into partially overlapped texture/picture
patches with the fixed patch size, and then these patches will be individually feeded into the subsequent fusion network, avoiding the resizing operation induced distortions.
In summary, we list the salient contributions as follows:

$\bullet$ We propose to conduct full reference quality assessment for
SCIs in a multi-level manner, which is perceptually consistent with
the real multi-level HSV;

$\bullet$ We propose a novel sparse representation based metric to
measure structure similarity globally, which can effectively
complement to the conventional local measurement for a trustworthy
quality assessment;

$\bullet$ We design a novel selective deep fusion network to integrate
the local and global structure similarity towards a high-level
processing and understanding, giving rise to much better accuracy than
the conventional hand-crafted solutions;

$\bullet$ We newly devise a novel content-aware decomposition scheme to
further improve the robustness of our SCI quality assessment;

$\bullet$ The source code and results are publicly available at \underline{\url{https://github.com/HongmengZhao/SR-CNN}}.

\section{related work}
Most of the existing FR (Full Reference) IQA methods are designed for natural scene images (NSIs), which frequently utilize pixel-wise measurements to evaluate the differences between the reference image and its distorted version.
Thus, the sensed differences, which is positively related to the distortion level, can be used to predict the overall quality score.

\subsection{Structure Similarity Metrics}
As a milestone FR metric for NSIs, Zhou et al.~\cite{Zhou2004Image} conducts IQA mainly from the structural similarity (SSIM) perspective, which is inspired from the phenomenon that the HVS is highly sensitive to structural information, and multiple improved subsequent works are proposed latter~\cite{Zhou2011Information,Zhang2013Edge,Wufeng2014Gradient}.
Different from the quality assessment of NSIs, the existence of various image content types in SCIs (e.g. computer generated textures and graphics) makes the quality assessment more challenging.
To appropriately reveal the correlation between the SSIM captured artifacts and the overall quality scores, Gu et al.~\cite{Gu2015Screen} proposes to re-weight the classic SSIM~\cite{Zhou2004Image} with structural degradation measurement computed using SSIM on the original SCI and its distorted version via a low-pass filter.
Moreover, since the HVS may pay more attention to salient object~\cite{chen15sod,chen2017video,chen2018bi,CC2019TMM2}, Gu et al.~\cite{Gu2016Saliency} further integrates the structural degradation measurement with structure variation based saliency clue to improve the overall quality score.
To better perceive structural degradation, Ni et al.~\cite{Ni2016Gradient} proposes to utilize both gradient direction and gradient magnitude to yield the overall quality assessment by employing a deviation-based pooling strategy, which is further improved by adopting the weighted pooling scheme to fuse the edge contrast degree with edge width variation~\cite{Ni2016Screen}.
Meanwhile, the above-mentioned edge information can also be expanded to simultaneously consider edge contrast, edge width and edge direction~\cite{Ni2017ESIM}.
Since the Gabor filters are highly consistent with the response of the HVS, Ni et al.~\cite{Ni2018A} resorted the Gabor filter to measure the global similarity between the reference image and its distorted version, achieving much improved quality assessment performance.

\subsection{Content-aware Fusion}

Despite the above-mentioned efforts to investigate low-level venues in accordance to the real HVS, these approaches may not be able to perform well occasionally due to its less consideration of the content types of the given SCIs.
Thus, Yang et al.~\cite{Huan2015Perceptual} proposes to separately measure sharpness similarity for image regions with different content types, i.e., texture regions and picture regions.
Similarly, Wang et al.~\cite{Wang2016Objective} proposed to incorporate both visual field adaptation and information content weighting into structural similarity based local quality assessment by adopting the information content model~\cite{Wang2007Spatial}.
Since the distortion metrics are computed individually on different content types, Fang et al.~\cite{Fang2017Objective} proposes to use the uncertainty weighting scheme to fuse the visual quality of textual and pictorial regions effectively, achieving remarkable performance improvement.
Although the above mentioned works have improved the quality assessment performance significantly, its hand-crafted nature has a tendency to reach performance bottle neck.
\begin{figure*}[t]
\centering
\includegraphics[width=1\linewidth]{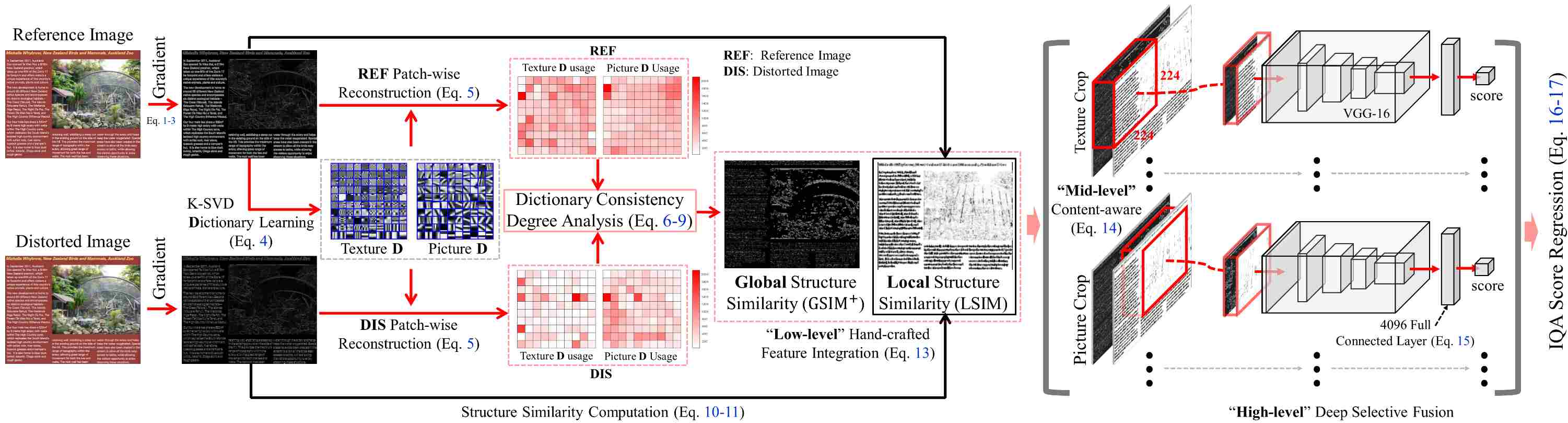}
\caption{The overall pipeline of our proposed multi-level SR-CNN method. The left part shows the ``low-level'' local/global structure-aware similarity measurement (Sec.~\ref{sec:Features}), and the right part demonstrates the ``mid-level'' content-aware patch selection and the ``high-level'' selective deep fusion toward the optimal quality score regression (Sec.~\ref{sec:fusion}).}
\label{fig:pipeline}
\end{figure*}

\subsection{Deep Learning based Methods}
After entering the deep learning era, multiple convolutional neural networks (CNNs) based methods have been developed to solve the IQA problem for SCIs.
For instance, Zuo et al.~\cite{Zuo2016Screen} proposes to directly input the sequentially cropped patches into CNNs to automatically reveal deep features for its subsequent quality score regression.
To shrink the regression problem domain, Cheng et al.~\cite{Cheng2018A} adopts a predefined saliency threshold to assign lower weight to the relatively less salient patches.
Different form pervious works which solely divide SCIs into texture and picture regions, Zhang et al.~\cite{Zhang2018Quality} proposes to employ CNNs-based content classifier to assign image regions into a novel SCI content, i.e., computer-graphics/cartoons.
Although this newly introduced content type is actually in line with the real HVS, it still follows the conventional hand-crafted manner to conduct its subsequent quality assessment with limited accuracy.
Also, the NSIs based deep models can be adapted for SCIs by using the off-the-shelf naturalization deep network, inducing significant performance improvement~\cite{Jiang2019DeepOM}, in which Chen et al.~\cite{Chen2018Naturalization} uses the naturalization module to transform IQA of SCIs into IQA of NSIs.
Most recently, Jiang et al.~\cite{Jiang2019DeepOM} proposes a quadratic optimized model to select representative image patches from the pre-trained deep model, and thus the quality assessment results predicted by this method are relatively more close to the DMOS (Differential Mean Opinion Score).

In summary, compared to the hand-crafted methods, the above mentioned deep learning based works have achieved significant performance improvements,
however, almost all these methods still follow the single-level or bi-level fashions, contradicting to the real HVS using multi-level clues in practice.
Thus, all the mentioned above motivates us to use deep learning framework to selectively integrate multi-level metrics for a high performance SCI quality assessment.

\section{Method Overview}
As shown in Fig.~\ref{fig:pipeline}, our method mainly consists of two stages: 1) Structure-aware global feature computation; 2) Global and local deep selective fusion.
Given a reference SCI (screen content image), we propose to conduct patch-wise dictionary learning to obtain multiple ``structure-aware'' non-local basis.
To achieve it, we conduct the accelerated K-SVD~\cite{aharon2006k} algorithm over the reference gradient image to formulate a 100-codeword dictionary respectively for texture (see Texture D in Fig.~\ref{fig:pipeline}) and picture regions (see Picture D in Fig.~\ref{fig:pipeline}).
Since the learned dictionaries can globally represent the consistent non-local structures of the reference image, we propose to globally predict the distortion degree of the given input image by measuring the differences between its codeword usage and the codeword usage in its reference image.
Meanwhile, we also adopt the local structure similarity measurement to complement the above-mentioned global one.
In particular, we adopt the newly designed selective deep fusion network for an optimal fusion state between these two measurements, achieving accurate IQA predictions eventually.

\section{Structure-aware Similarity Features Computation}
\label{sec:Features}
Given a reference image (REF) and its distorted version (DIS), we first formulate its non-local basis (codewords) by conducting the K-SVD learning procedure.
Next, from the codeword usage perspective, we measure the differences between REF and DIS to represent our structure-aware global similarity.

\subsection{Preliminaries}
In general, for any screen content image (\textbf{I}), its structural information can be represented by using its two directional gradient maps ($G(\textbf{I})$), which can be formulated as Eq.~\ref{eq:gradient}.
\begin{equation}
G(\textbf{I})= \sqrt{(h_x*\textbf{I})^2+(h_y*\textbf{I})^2},
\label{eq:gradient}
\end{equation}
where $h_x = [-\frac{1}{2}\ 0\ \frac{1}{2}]$ and $h_y = [-\frac{1}{2}\ 0\ \frac{1}{2}]^{\rm T}$ denote the vertical kernel and horizonal kernel respectively, and the operator $*$ denotes the convolution operation.
So, given a pair of reference image ($\textbf{I}_r$) and its distorted version ($\textbf{I}_d$), we can represent their gradient maps respectively as $G(\textbf{I}_r)$ and $G(\textbf{I}_d)$.

To obtain the global structure-aware dictionary of the reference image, we first adopt the patch-wise decomposition (patch size: $8\times 8$, overlapping rate: 50\%) over $\textbf{I}_r$ and then learn the common structure basis from these segmented image patches.
To balance the trade-off between computational cost and performance, we randomly select maximum 20,000 texture patches and maximum 20,000 picture patches to respectively formulate the texture dictionary (with 100 codewords) and picture dictionary (with 100 codewords), in which we adopt the off-the-shelf method~\cite{Jaderberg2014Deep} to separate texture patches from picture patches.

Here we represent the selected patch-wise texture/picture data respectively as Eq.~\ref{eq:YRT}/Eq.~\ref{eq:YRP}.
\begin{equation}
\textbf{Y}_{rt}=\{\textbf{t}_1,\textbf{t}_2,...,\textbf{t}_{TN}\}\in\mathbb{R}^{64\times TN},
\label{eq:YRT}
\end{equation}
\begin{equation}
\textbf{Y}_{rp}=\{\textbf{p}_1,\textbf{p}_2,...,\textbf{p}_{PN}\}\in\mathbb{R}^{64\times PN},
\label{eq:YRP}
\end{equation}
where $t$ and $p$ respectively denote the columnized texture patches and the columnized picture patches, $TN\leq 20,000$ and $PN\leq 20,000$ respectively denote the total number of the selected texture/picture patch numbers in practice.
Thus, the dictionaries learning procedure can be formulated as Eq.~\ref{eq:KSVD}.
\begin{equation}
\min\limits_{\textbf{D}_{t/p},\textbf{X}_{t/p}} \parallel \textbf{Y}_{t/p}-\textbf{D}_{t/p}\cdot\textbf{X}_{t/p} \parallel_F^2, \ \ \ s.t.\ \ \forall i,\parallel \textbf{x}_i\parallel_0 \leq \rm T_0,
\label{eq:KSVD}
\end{equation}
where $\textbf{D}_t$ and $\textbf{D}_p$ respectively denote the texture dictionary and the picture dictionary with total 100 dictionary atoms ($\textbf{D}_{t/p}\in\mathbb{R}^{64\times 100}$), $\textbf{X}=\{\textbf{x}_1,\textbf{x}_2,...,\textbf{x}_{TN or PN}\}\in\mathbb{R}^{100\times (TN\ or\ PN)}$ is a sparse matrix representing the dictionary usage toward the input \textbf{Y}.

\begin{figure}[t]
\centering
\includegraphics[width=0.6\linewidth]{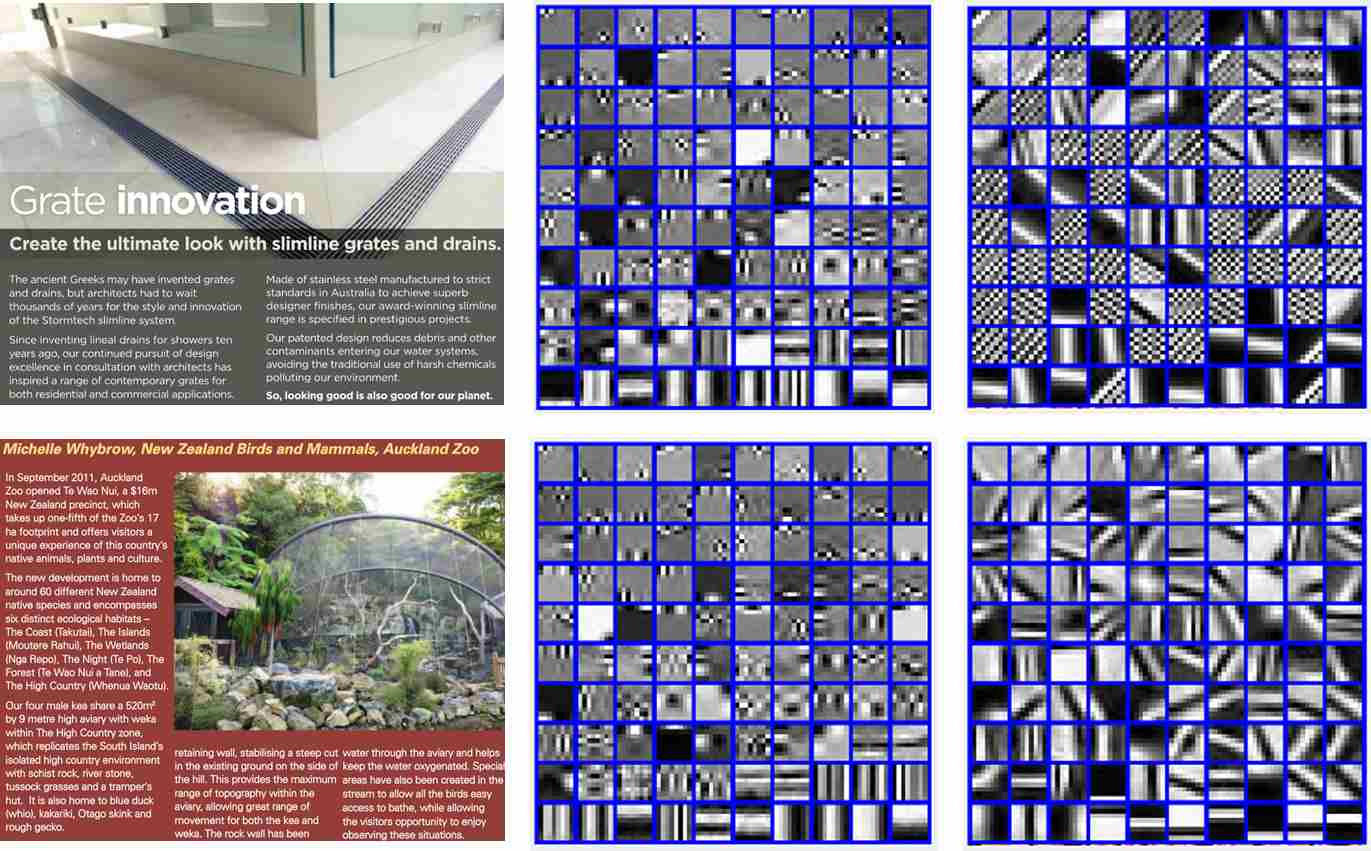}
\caption{An illustration of the revealed textual/picture dictionaries. [Column 1]: reference SCIs. [Column 2]: the corresponding texture dictionaries. [Column 3]: the corresponding picture dictionaries.}
\label{fig:DDemo}
\end{figure}

Actually, Eq.~\ref{eq:KSVD} is a common thread of the sparse representation problem~\cite{CC2019CVPR,CC2020INS}, where the left Frobenius norm ensures a minimization of the reconstructed mean standard error while the right part constraint maintain a sparse state toward the dictionary atom usage.
Thus, Eq.~\ref{eq:KSVD} can be alternative solved, i.e., we utilize SVD decomposition to iteratively estimate \textbf{D}, and then the code matrix \textbf{X} can be obtained by using the orthogonal matching pursuit (OMP~\cite{pati1993orthogonal}) over the current generated dictionary \textbf{D}.
In this way, we can easily obtain a complete high-quality dictionary and its corresponding sparse coding.
Also, we demonstrate the difference between the learned texture dictionary ($\textbf{D}_t$) and picture dictionary ($\textbf{D}_p$) in Fig.~\ref{fig:DDemo}.

\subsection{Multi-level Structure-aware Similarity}
\indent
Since the SCIs (screen content images) frequently exhibit sharp boundaries, the conventional ``local'' structure similarity has been widely adopted to measure the distortion degree of the input image.
However, such ``local'' protocol may contradict with the real HVS (human visual system) in the following two aspects:\\
1) The HVS may easily overlook the structure related tiny distortions in cluttered backgrounds;\\
2) Moreover, our HVS tends to sweep though the entire screen while assigning an objective SCI score.

To satisfy the former attribute 1), we propose to limit the structure similarity measurement scope within the common basis $\textbf{D}_{t/p}$ which are previously obtained from the reference SCI via Eq.~\ref{eq:KSVD}.
Thus, the tiny structure distortions can be automatically filtered during the IQA process.
Meanwhile, as for the attribute 2), we propose to conduct IQA by measuring the structure similarity from the ``global'' perspective, in which we reconstruct the distorted image over $\textbf{D}_{t/p}$, and then measure the dictionary usage consistency degree between REF (reference image) and DIS (distorted image) as the predicted image quality degree.

Suppose we have already obtained the reference texture dictionary ($\textbf{D}_t^r$) and the reference picture dictionary ($\textbf{D}_p^r$), the reconstruction procedure toward the distorted image can be formulated as Eq.~\ref{eq:Reconstruction}.
\begin{equation}
\min\limits_{\textbf{X}_{t/p}^{d}} \parallel \textbf{Y}_{t/p}^d-\textbf{D}_{t/p}^r\cdot\textbf{X}_{t/p}^d \parallel_F^2, \ \ \ s.t.\ \ \forall i,\parallel \textbf{x}_i\parallel_0 \leq \rm T_0,
\label{eq:Reconstruction}
\end{equation}
where the superscript $r$ and $d$ denote the reference image and the distorted image, respectively.
We also have adopted the~\cite{Zhou2018Local} proposed $k$ nearest clustering strategy to accelerate the reconstruction procedure.
By reconstructing the distorted image over the structure dictionary provided by reference (Eq.~\ref{eq:KSVD}), we can obtain the sparse coding of the distorted image, i.e., the distorted texture coding ($\textbf{X}_{t}^{d}$) and the distorted picture coding ($\textbf{X}_{p}^{d}$).
Actually, the sparse coding of the distorted image is partially different to the sparse coding of the reference image, which can well perceive the distinction between reference and distorted SCIs.
Thus, we utilize the dictionary atom usage overlapping rate to measure the distortion degree, in which both the $\textbf{X}_{t/p}^{d}$ and the $\textbf{X}_{t/p}^{r}$ are negatively related to the distortion degree.
Here we formulate such computation process as Eq.~\ref{eq:OverLapping}.
\begin{equation}
\begin{aligned}
\rm{GSIM}_\emph{t/p}(\emph{i})&=\frac{\parallel\emph{col}(\textbf{X}_{\emph{t/p}}^{\emph{d}}\cdot\textbf{X}_{\emph{t/p}}^{\emph{r}},\emph{i})\parallel_0}{\parallel \emph{col}(\textbf{X}_{\emph{t/p}}^{\emph{d}},\emph{i})\parallel_0+\parallel\emph{col}(\textbf{X}_{\emph{t/p}}^{\emph{r}},\emph{i})\parallel_0}\\
&=\frac{\parallel \emph{col}(
{\left[
 \begin{smallmatrix}
 \textbf{x}_{\emph{1,1}}^{\emph{d}}            & \cdots & \textbf{x}_{\emph{1,TN/PN}}^{\emph{d}}      \\
\vdots &\vdots  &\vdots \\
 \textbf{x}_{\emph{100,1}}^{\emph{d}}             & \cdots & \textbf{x}_{\emph{100,TN/PN}}^{\emph{d}}       \\
\end{smallmatrix}
\right]}
\cdot
{\left[
 \begin{smallmatrix}
 \textbf{x}_{\emph{1,1}}^{\emph{r}}            & \cdots & \textbf{x}_{\emph{1,TN/PN}}^{\emph{r}}      \\
 \vdots & \vdots  & \vdots \\
 \textbf{x}_{\emph{100,1}}^{\emph{r}}             & \cdots & \textbf{x}_{\emph{100,TN/PN}}^{\emph{r}}       \\
 \end{smallmatrix}
\right]},\emph{i})
\parallel_0}{\parallel \emph{col}({\left[
 \begin{smallmatrix}
 \textbf{x}_{\emph{1,1}}^{\emph{d}}            & \cdots & \textbf{x}_{\emph{1,TN/PN}}^{\emph{d}}      \\
 \vdots & \vdots  & \vdots \\
 \textbf{x}_{\emph{100,1}}^{\emph{d}}             & \cdots & \textbf{x}_{\emph{100,TN/PN}}^{\emph{d}}       \\
 \end{smallmatrix}
\right]},\emph{i})\parallel_0+\parallel \emph{col}({\left[
 \begin{smallmatrix}
 \textbf{x}_{\emph{1,1}}^{\emph{r}}            & \cdots & \textbf{x}_{\emph{1,TN/PN}}^{\emph{r}}      \\
 \vdots & \vdots  & \vdots \\
 \textbf{x}_{\emph{100,1}}^{\emph{r}}             & \cdots & \textbf{x}_{\emph{100,TN/PN}}^{\emph{r}}       \\
\end{smallmatrix}
\right]},\emph{i})\parallel_0}
\end{aligned}
\label{eq:OverLapping}
\end{equation}
where the global texture structure similarity $\rm{GSIM}_\emph{t}\in\mathbb{R}^{1\times \emph{TN}}$, and the picture structure similarity $\rm{GSIM}_\emph{p}\in\mathbb{R}^{1\times \emph{PN}}$, $TN$ and $PN$ respectively denote the number of selected texture and picture patches, function $col(\textbf{X},i)$ returns the $i$-th column of its input matrix \textbf{X}, $\parallel\cdot\parallel_0$ denotes the $l_0$-norm.

So far, the computed GSIM (Eq.~\ref{eq:OverLapping}) can well measure the global structure similarity from the perspective of sparse reconstruction.
Meanwhile, to further measure the structure similarity from the dictionary perspective (further strengthen the global attribute), we propose to introduce the dictionary atom usage coefficients into Eq.~\ref{eq:OverLapping} as its patch-wise weights, because the dictionary atoms with large coefficient variations should be considered more during the above-mentioned measurement, and vice versa.
Here we formulate the patch-wise dictionary atom weight as Eq.~\ref{eq:CW}.
\begin{equation}
\begin{split}
\rm{CW}_\emph{t/p}(\emph{i})=\underset{overlapped\ atom\ mask} {\underbrace{\lceil\frac{\emph{col}(\textbf{X}_{\emph{t/p}}^{\emph{d}}\cdot\textbf{X}_{\emph{t/p}}^{\emph{r}},\emph{i})}{\max\{\emph{col}(\textbf{X}_{\emph{t/p}}^{\emph{d}}\cdot\textbf{X}_{\emph{t/p}}^{\emph{r}},\emph{i})\}}\rceil}}
\cdot
\underset{coefficient\ change}{\underbrace{\frac{|\textbf{C}_{\emph{t/p}}^{\emph{d}}-\textbf{C}_{\emph{t/p}}^{\emph{r}}|}{|\textbf{C}_{\emph{t/p}}^{\emph{d}}|+|\textbf{C}_{\emph{t/p}}^{\emph{r}}|}}},
\label{eq:CW}
\end{split}
\end{equation}
In Eq.~\ref{eq:CW}, $\lceil\cdot\rceil$ denotes the ceiling operation; The left part $\in\{0,1\}$ indicates the overlapped dictionary atom usage toward the $i$-th image patch, while the right part measures the difference of the dictionary usage coefficient (\textbf{C}), which can be formulated by Eq.~\ref{eq:AtomFrequency}.
\begin{equation}
\textbf{C}_{\emph{t}}^{\emph{d/r}} = \rm \frac{\sum(\textbf{X}_{\emph{t}}^{\emph{d/r}})}{\emph{TN}}, \ \ \ \ \textbf{C}_{\emph{p}}^{\emph{d/r}} = \rm \frac{\sum(\textbf{X}_{\emph{p}}^{\emph{d/r}})}{\emph{PN}},
\label{eq:AtomFrequency}
\end{equation}
where the superscript $d/r$ denotes the reference image or the distorted image, $\textbf{C}\in\mathbb{R}^{1\times {TN\ or\ PN}}$ denotes the dictionary atom usage variation, and function $\sum$ represents the column-wise summation.

At this point, we can re-formulate Eq.~\ref{eq:OverLapping} into the following Eq.~\ref{eq:GSIMCW}.
\begin{equation}
\rm{GSIM}^+_{t/p}(i) = \{CW_\emph{t/p}(i)\}^{\alpha_1}\cdot \{\rm{GSIM}_\emph{t/p}(i)\}^{\beta_1},
\label{eq:GSIMCW}
\end{equation}
where $\alpha_1$ and $\beta_1$ are two predefined parameters to control the balance between the dictionary coefficient component and the dictionary overlapping component. In this paper, these two parameter values are empirically assigned to $\alpha_1=0.2$ and $\beta_1=1$.

\begin{figure}[t]
\centering
\includegraphics[width=0.6\linewidth]{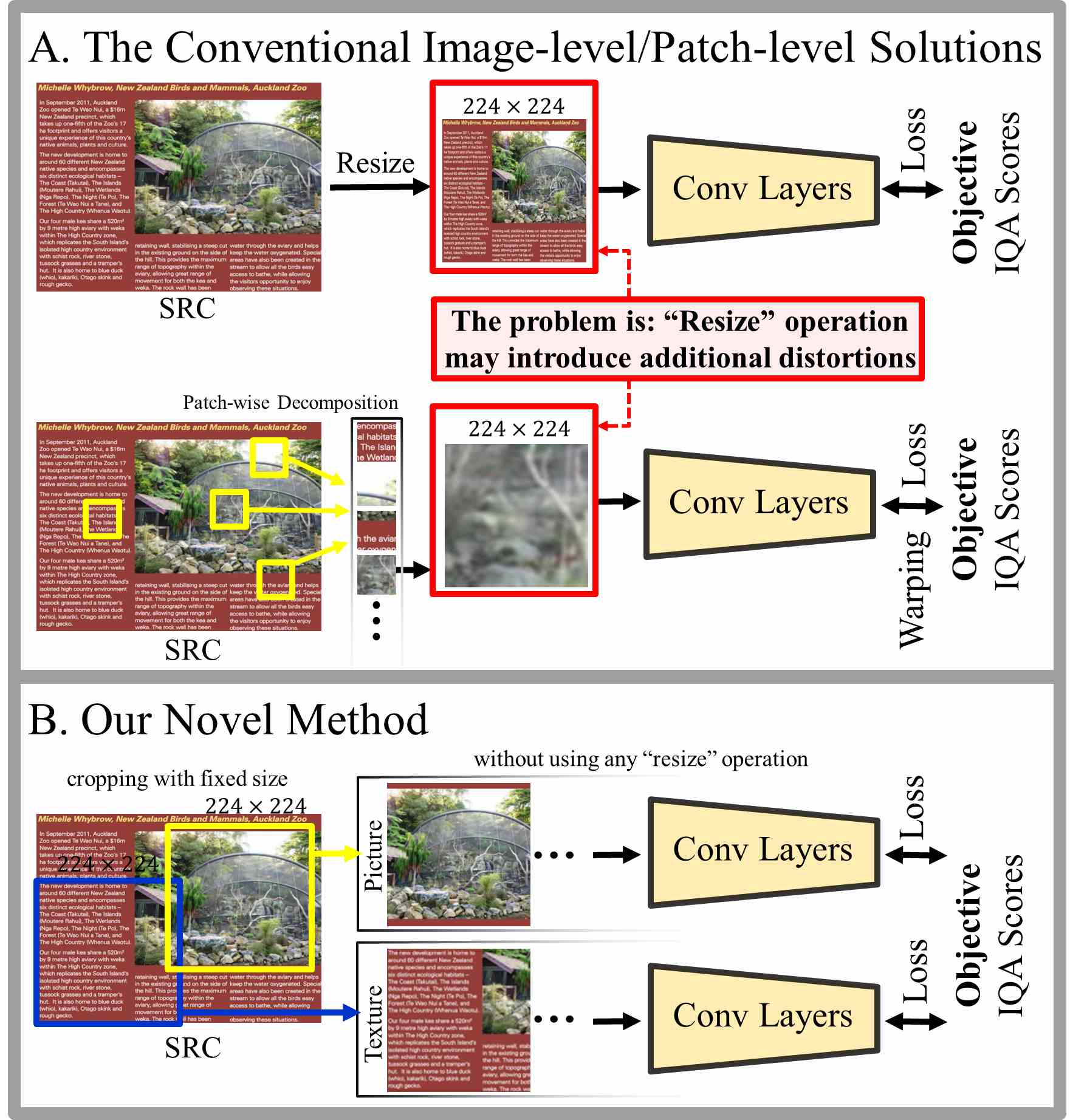}
\caption{The current main stream deep learning based methods~\cite{Jiang2019DeepOM,Chen2018Naturalization} frequently resort the ``resize'' operation to adapt patch-wise input data for their pre-trained feature backbones, which is inevitable to bring additional distortions, making the quality score regression difficult. So, we adopt the content-aware patch selection (Sec.~\ref{sec:PAS}) to solve this problem.}
\label{fig:CNN}
\end{figure}

Meanwhile, though the $\rm{GSIM}^+_{t/p}$ can well represent the global structure similarity already, yet the local structure similarity should also be considered, because the HVS may also be easily attracted to the conspicuous tiny local structure artifacts in plain texture regions.
Therefore, inspired by the conventional structure similarity computations~\cite{Liu2012Image,Lin2014VSI,Xing2017A}, we propose to formulate our local structure similarity computation between the reference gradient map ($G(\textbf{I}^r)$, Eq.~\ref{eq:gradient}) and the distorted gradient map ($G(\textbf{I}^d)$, Eq.~\ref{eq:gradient}) within a patch-wise manner as Eq.~\ref{eq:LSIM}.
\begin{equation}
\begin{split}
\rm{LSIM}&(i) = \\
&\frac{\mu^r_i\cdot \mu^d_i+\rm{C}_1}{\{\mu^r_i\}^2+\{\mu^d_i\}^2+\rm{C}_1}\cdot\frac{\sigma^r_i\cdot \sigma^d_i+\rm{C}_2}{\{\sigma^r_i\}^2+\{\sigma^d_i\}^2+\rm{C}_2},
\end{split}
\label{eq:LSIM}
\end{equation}
where $\mu_i$ and $\sigma_i$ respectively denote the mean value and the standard value of the $i$-th gradient patch, which can be formulated by Eq.~\ref{eq:LSIMDetail}, $\rm C_1$ and $\rm C_2$ are two constant values to avoid the instability in the case that the denominator is close to zero, i.e., $\rm C1=(0.01*255)^2$ and $\rm C2=(0.03*255)^2$.
\begin{equation}
\mu^{r/d}_i = mean\{\emph{col}(\textbf{Y}^{r/d},i)\},\ \sigma^{r/d}_i = std\{\emph{col}(\textbf{Y}^{r/d},i)\}.
\label{eq:LSIMDetail}
\end{equation}

\begin{figure*}[t]
\centering
\includegraphics[width=1\linewidth]{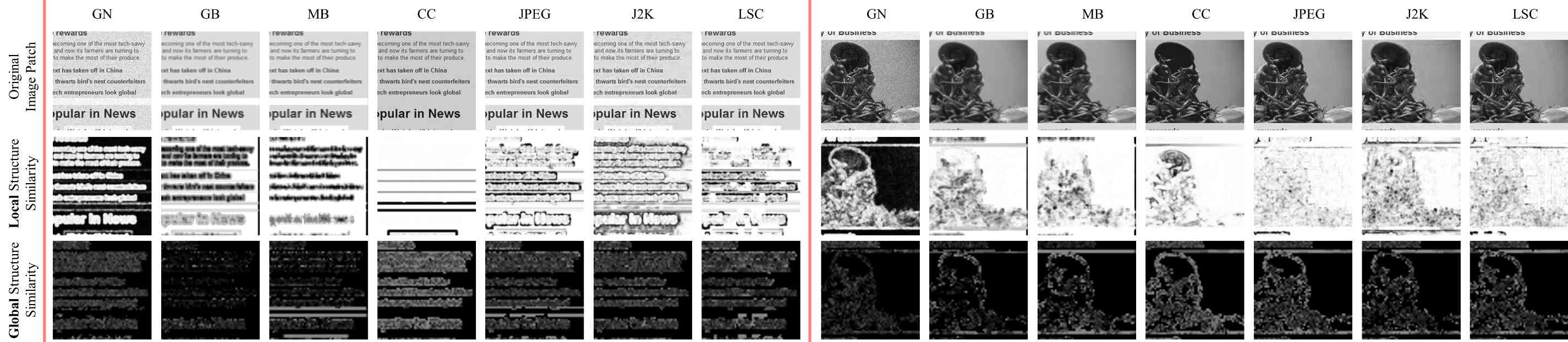}
\caption{The revealed low-level feature map demonstration. The left/right parts demonstrate several texture/picture cases. Columns from left to right respectively represent different distortion types, including GN (Gaussian Noise), GB (Gaussian Blur), MB (Motion Blur), CC (Contrast Change), JPEG (JPEG Compression), J2K (JPEG2000 Compression), LSC (Layer-segmentation based Compression). [Row 1]: original image patches; [row 2]: the corresponding Local Structure Similarity; [Row 3]: the corresponding Global Structure Similarity.}
\label{fig:feature}
\end{figure*}

The patch-wise quality assessment score (PS) can be computed by using multiplicative based fusion toward $\rm GSIM^+$ (Eq.~\ref{eq:GSIMCW}) and $\rm LSIM$ (Eq.~\ref{eq:LSIM}) as the following:
\begin{equation}
\rm{PS}(\emph{i}) = \left\{ \begin{array}{ll}
\{\rm{LSIM}(\emph{i})\}^{\alpha_2}\cdot \{\rm{GSIM}^+_\emph{t}(\emph{i})\}^{\beta_2}\ \ \ \emph{if}\ \emph{i}\in \textbf{Y}_\emph{t}\\
\{\rm{LSIM}(\emph{i})\}^{\alpha_2}\cdot \{\rm{GSIM}^+_\emph{p}(\emph{i})\}^{\beta_2}\ \ \ \emph{if}\ \emph{i}\in \textbf{Y}_\emph{p}\\
\end{array}\right.,
\label{eq:PatchScore}
\end{equation}
where $\alpha_2=0.6$ and $\beta_2=1$ are another two pre-defined weighting parameters to control the balance between the LSIM and $\rm GSIM^+$.
Then, the image level quality assessment score (FSIM) can be obtained via averaging all the above computed patch-wise quality assessment scores as Eq.~\ref{eq:ImageScore}.
\begin{equation}
\rm FSIM = \frac{1}{\emph{TN}}\sum_{\emph{i}\in\textbf{Y}_\emph{t}} PS_\emph{i} + \frac{1}{\emph{PN}}\sum_{\emph{j}\in\textbf{Y}_\emph{p}} PS_\emph{j}.
\label{eq:ImageScore}
\end{equation}
So far, the IQA score can be much improved by using Eq.~\ref{eq:ImageScore} to fuse both local and global low-level metrics (its pictorial demonstration can be found in Fig.~\ref{fig:feature}, and its quantitative proofs can be found in Tab.~\ref{tab:Components}).
However, it is difficult to use such handcrafted fusion (e.g., Eq.~\ref{eq:LSIM}) to realize an optimal complementary fusion status.
So, we will resort the newly devised selective deep fusion to perform automatical selective fusion between $\rm LSIM$ and $\rm GSIM^+$ for an optimal complementary fusion status, which will be detailed in the next section.

\section{Selective Deep Fusion}
\label{sec:fusion}
Given a reference SCI and a distorted SCI, we have already obtained the global structure similarity feature map ($\rm GSIM^+$) and the local structure similarity feature map (LSIM), and we propose to simultaneously feed these two feature maps into the selective deep fusion network for an optimal complementary fusion status.

\subsection{Content-aware Adaptive Mid-level Patch Decomposition}
\label{sec:PAS}
Given an off-the-shelf deep network (i.e., we simply adopt the vanilla VGG-16 in this paper), its input layer can only receive input data with fixed size, e.g., $224\times224$ for the VGG-16.
Thus, the conventional methods frequently resort the ``resize'' operation either over the entire image or the image patches to enable a valid input, see pictorial demonstrations in Fig.~\ref{fig:CNN}-A.

Though much progresses have been made by the previous deep learning based works, the widely adopted ``resize'' operation (e.g., image-level down-sampling or patch-wise up-sampling) may easily induce additional image distortions, leading to limited quality assessment performance eventually, please refer to the quantitative proofs in Tab.~\ref{tab:Components}.
To solve the above mentioned problem, here, we advocate a novel mid-level patch decomposition strategy, and its method pipeline can be found in Fig.~\ref{fig:CNN}-B.
Our novel strategy is capable of conserving both the local and global spatial information (ensured by using a large patch number $n$, see proofs in Tab.~\ref{tab:patches}), yet it can avoid the resizing operation induced patch distortions.

\begin{table*}[!t]
\centering
\caption{Performance comparisons (via 80\%-20\% training-testing partition) of the deep learning based SOTA SCI quality assessment models on the SIQAD and the SCID datasets. The column-wise bests are marked by bold font. The ``-'' denotes the absent data.}
   \vspace{-0.2cm}
\resizebox{1\linewidth}{!}{
    \begin{tabular}{ccccccccc|ccccccccc}
    \toprule
    \toprule
          & \multicolumn{8}{c|}{\textbf{SIQAD Dataset}}                                    & \multicolumn{9}{c}{\textbf{SCID Dataset}} \\
    Distortion & GN    & GB    & MB    & CC    & JPEG  & J2K   & LSC   & \textbf{Overall} & GN    & GB    & MB    & CC    & JPEG  & J2K   & HEV. & CQD   & \textbf{Overall} \\
    \midrule
          & \multicolumn{8}{c|}{PLCC}                                     & \multicolumn{9}{c}{PLCC} \\
    CNN-SQE18~\cite{Zhang2018Quality} & -     & -     & -     & -     & -     & -     & -     & 0.904  & -     & -     & -     & -     & -     & -     & -     & -     & 0.915  \\
    QODCNN-FR19~\cite{Jiang2019DeepOM} & 0.918  & 0.934  & 0.907  & \textbf{0.866} & 0.848  & 0.857  & \textbf{0.897} & 0.914  & -     & -     & -     & -     & -     & -     & -     & -     & - \\
    RIQA19~\cite{jiang2019no} & \textbf{0.920} & 0.930  & 0.905  & 0.857  & \textbf{0.872} & \textbf{0.891} & 0.857  & 0.911  & -     & -     & -     & -     & -     & -     & -     & -     & - \\
    SIQA-DF-II19~\cite{jiang2019screen} & 0.912  & 0.924  & 0.890  & 0.844  & 0.829  & 0.828  & 0.858  & 0.900  & -     & -     & -     & -     & -     & -     & -     & -     & - \\
    SR-CNN & 0.916  & \textbf{0.935} & \textbf{0.912} & 0.852  & 0.866  & 0.872  & 0.875  & \textbf{0.916} & \textbf{0.967} & \textbf{0.938} & \textbf{0.938} & \textbf{0.877} & \textbf{0.963} & \textbf{0.961} & \textbf{0.918} & \textbf{0.922} & \textbf{0.939} \\
          & \multicolumn{8}{c|}{SROCC}                                    & \multicolumn{9}{c}{SROCC} \\
    CNN-SQE18~\cite{Zhang2018Quality} & 0.893  & 0.924  & 0.904  & 0.665  & 0.847  & 0.862  & 0.887  & 0.894  & 0.949  & 0.907  & 0.878  & 0.745  & 0.947  & 0.939  & \textbf{0.898} & \textbf{0.904} & 0.914  \\
    QODCNN-FR19~\cite{Jiang2019DeepOM} & \textbf{0.907} & 0.921  & 0.895  & \textbf{0.778} & 0.829  & 0.835  & \textbf{0.898} & 0.907  & -     & -     & -     & -     & -     & -     & -     & -     & - \\
    RIQA19~\cite{jiang2019no} & 0.901  & \textbf{0.936} & 0.893  & 0.714  & 0.859  & \textbf{0.890} & 0.883  & 0.900  & -     & -     & -     & -     & -     & -     & -     & -     & - \\
    SIQA-DF-II19~\cite{jiang2019screen} & 0.901  & 0.910  & 0.880  & 0.728  & 0.812  & 0.816  & 0.858  & 0.888  & -     & -     & -     & -     & -     & -     & -     & -     & - \\
    SR-CNN & 0.896  & 0.928  & \textbf{0.910} & 0.748  & \textbf{0.860} & 0.874  & 0.875  & \textbf{0.908} & \textbf{0.957} & \textbf{0.935} & \textbf{0.924} & \textbf{0.810} & \textbf{0.957} & \textbf{0.940} & 0.885  & 0.896  & \textbf{0.940} \\
          & \multicolumn{8}{c|}{RMSE}                                     & \multicolumn{9}{c}{RMSE} \\
    CNN-SQE18~\cite{Zhang2018Quality} & -     & -     & -     & -     & -     & -     & -     & 6.115  & -     & -     & -     & -     & -     & -     & -     & -     & 5.761  \\
    QODCNN-FR19~\cite{Jiang2019DeepOM} & 5.963  & 5.454  & 5.251  & 6.381  & 5.141  & 5.286  & \textbf{3.857} & 5.801  & -     & -     & -     & -     & -     & -     & -     & -     & - \\
    RIQA19~\cite{jiang2019no} & 6.172  & 5.712  & 5.283  & 6.617  & 4.767  & \textbf{4.597} & 4.476  & 5.880  & -     & -     & -     & -     & -     & -     & -     & -     & - \\
    SIQA-DF-II19~\cite{jiang2019screen} & 6.115  & 5.768  & 5.791  & 6.747  & 5.840  & 5.812  & 4.462  & 6.242  & -     & -     & -     & -     & -     & -     & -     & -     & - \\
    SR-CNN & \textbf{5.733} & \textbf{5.089} & \textbf{5.131} & \textbf{6.205} & \textbf{4.496} & 4.856  & 3.915  & \textbf{5.683} & \textbf{3.127} & \textbf{3.558} & \textbf{3.679} & \textbf{4.152} & \textbf{3.966} & \textbf{4.225} & \textbf{ 5.249} & \textbf{4.856} & \textbf{4.830} \\
    \bottomrule
    \bottomrule
     \end{tabular}}%
  \label{tab:overall}%
\end{table*}%

To be specific, we adaptively select multiple mid-level patches with fixed size $m$ (e.g., $m=224$ for the VGG-16 network), and then feed these mid-level patches related feature maps (i.e., $\rm GSIM^+$ and LSIM) to the deep selective fusion network.
Specially, we have assigned an identical score (i.e., image-level objective score)for each of these mid-level patches due to the following two aspects:\\
1) We have adopted a relatively large patch size, i.e., $224\times 224$, and such size is much larger than the common-thread small-size patch-wise partitions (e.g., $11\times11$), which we believe our patch-wise partition should be categorized as a non-local manner, and thus it may be reasonable to use image-level scores directly;\\
2) Instead of using RGB patches as input, our mid-level patch-wise partition is performed on the feature maps (i.e., LSIM and $\rm GSIM^+$), in which the $\rm GSIM^+$ can provide substantial global information even in the case of using such patch-wise formulation.\\
Due to the different intrinsic attributes for content types, the deep fusion process itself should also treat these patches discriminatively.
Thus, given a reference SCI, we conduct K-means clustering using HOG (i.e., $8\times 8$ overlapped tiny patches) respectively over the texture regions ($\textbf{I}^r_t$) and the picture regions ($\textbf{I}^r_p$).
Hence, it is intuitive to use the cluster center to coarsely locate our mid-level image patches, which can be formulated by Eq.~\ref{eq:MidP}.
\begin{equation}
(\xi_{x_i},\xi_{y_i})\gets ({\rm max}\{\xi_{x_i},\lfloor\frac{m-1}{2}\rfloor\}, {\rm max}\{\xi_{y_i},\lfloor\frac{m-1}{2}\rfloor\}),
\label{eq:MidP}
\end{equation}
where $\xi_{x_i}$ and $\xi_{y_i}$ respectively denote the cluster center coordinates ($x$-Horizontal, $y$-Vertical) of the $i$-th clustering, $\lfloor\cdot\rfloor$ denotes the floor operation.

Also, it is worthy mentioning that we choose to use the HOG for clustering due to the following three reasons:\\
1) In general, there frequently exists remarkable differences in appearance between texture and picture regions, however, it may be more easily to notice such differences from the gradient perspective, i.e., the gradient patterns in texture regions are frequently more regular than those in picture regions; \\
2) The HOG feature has a strong capability of representing gradient patterns with a relatively low computational cost.
In our implementation, we have conducted the HOG feature based clustering in the reference image;\\
3) Those texture/picture regions frequently share similar HOG patterns if these regions are similar in non-local appearances, ensuring the HOG based clustering centers can be scattered with low overlapping rate.
Meanwhile, it should be noted that we utilize the commonly used padding strategy to handle the cases (less than 5\% in our adopted datasets) that the SCI image size is smaller than the fixed mid-level patch size.
After $2\times n$ mid-level decomposition (i.e., $n$ texture patches and $n$ picture patches), we formulate these mid-level patches into $2\times n$ input data for a single SCI (channel=2), which will be further detailed in the next section.

\subsection{Quality Score Regression}
So far, we have sparsely represented the input SCI into multiple texture/picture patches with fixed size.
Since we aim to achieve an optimal complementary status between the LSIM and the $\rm GSIM^+$, for each mid-level patch, we propose to feed its LSIM and $\rm GSIM^+$ into the pre-trained deep network to compute deep feature as Eq.~\ref{eq:SFeat}.
\begin{equation}
\rm{SFeat}\in \mathbb{R}^{1\times 4096} = \rm{FNet}(\rm{LSIM}\oplus\rm{GSIM^+}),
\label{eq:SFeat}
\end{equation}
where the FNet denotes the deep feature computation subnet (i.e., we use the off-the-shelf VGG-16~\cite{Simonyan2014Very}), and its input is $224\times 224\times 2$, where $\oplus$ denotes the feature concatenation operation.

By using Eq.~\ref{eq:SFeat}, we obtain 4096 dimension deep feature, and then feed it into the regression subnet (RNet), which consists of 2 full connected layers, using typical Euclidean loss.
Also, as we have mentioned before that the textual and pictorial regions in the SCIs have different properties, we should train two regression subnets respectively, i.e., one for the textual patches ($\rm RNet_\emph{t}$), and another for the picture patches ($\rm RNet_\emph{p}$), and the overall network architecture can be found in the right part of Fig.~\ref{fig:pipeline}.
After network training, we formulate the overall quality assessment (FS) as Eq.~\ref{eq:FinalScore}.
\begin{equation}
\rm FS = \frac{0.5}{\emph{n}}\sum_{\emph{i}=1}^\emph{n} \rm RNet_\emph{t}\{SFeat_\emph{t}(\emph{i})\} + \frac{0.5}{\emph{n}}\sum_{\emph{i}=1}^\emph{n} RNet_\emph{p}\{SFeat_\emph{p}(\emph{i})\},
\label{eq:FinalScore}
\end{equation}
where the meaning of $n$ is identical to Sec.~\ref{sec:PAS}, and its ablation study can be found in Tab.~\ref{tab:patches}.

Meanwhile, because FS (Eq.~\ref{eq:FinalScore}) is mainly derived from the gradient based structure similarity, it it intuitive to use the pixel-wise texture difference to benefit our method, thus we further use the multiplicative based fusion to integrate the Gabor filter based texture similarity~\cite{Ni2018A} into our estimated quality score to robust its performance as Eq.~\ref{eq:UpdatedFinalScore}.
\begin{equation}
\rm FS\gets FS\cdot GF,
\label{eq:UpdatedFinalScore}
\end{equation}
where GF is the off-the-shelf texture similarity degree provided by the work~\cite{Ni2018A}.

We implement our model using Matlab R2016a with Caffe~\cite{Jia2014Caffe} on a machine with NVIDIA GTX 1080Ti GPU, Intel Xeon W-2133 CPU and 32G RAM. The stochastic gradient descent (SGD) is used to optimise the network with an initial learning rate of $10^{\substack -5}$. The weight decay is $0.0005$, and the momentum is $0.9$. Our network is pre-trained on ImageNet~\cite{Deng2009ImageNet}. During testing, the average predicted scores of $2\times (n-1)$ patches (we have removed the largest and lowest scores empirically) as the final whole-image quality score, and we use ``SR-CNN'' to denote our method.
\begin{table*}[!t]
  \centering
  \caption{Cross-dataset quantitative comparisons of our proposed model and other SOTA methods on SIQAD dataset. The row-wise bests are marked with red color, the 2nd-bests are marked with blue color, and the 3rd-bests are marked with black bold font. The ``-'' denotes the absent data.}
     \vspace{-0.2cm}
  \resizebox{\linewidth}{!}{
    \begin{tabular}{ccccccccccccccc}
    \toprule
    \toprule
          & \multirow{2}[1]{*}{Distortion} & PieAPP~\cite{Prashnani2018PieAPP} & SFUW~\cite{Fang2017Objective}  & ESIM~\cite{Ni2017ESIM}  & CNN-SQE~\cite{Zhang2018Quality} & SVQI~\cite{Gu2017Evaluating}  & GFM~\cite{Ni2018A}   & MDOGS~\cite{fu2018screen} & EFGD~\cite{2019WangScreen}  & HRFF~\cite{2019Hybrid}  & QODCNN-FR~\cite{Jiang2019DeepOM} & RIQA~\cite{jiang2019no}  & SIQA-DF-II~\cite{jiang2019screen} & SR-CNN  \\
          &       & 2018  & 2017  & 2017  & 2018  & 2018  & 2018  & 2018  & 2019  & 2019  & 2019  & 2019  & 2019  &  \\
    \midrule
    \multirow{7}[1]{*}{\begin{sideways}PLCC\end{sideways}} & GN    & 0.898  & 0.887  & 0.889  & 　-    & \textbf{0.903} & 0.899  & 0.898  & 0.876  & 0.902  & 　-    & \textcolor[rgb]{.31,  .506,  .741}{\textbf{0.915}} & 　-    & \textcolor[rgb]{1,  0,  0}{\textbf{0.922}} \\
          & GB    & 0.840  & 0.923  & 0.923  & 　-    & 0.913  & 0.914  & 0.920  & \textbf{0.932} & 0.890  & 　-    & \textcolor[rgb]{.31,  .506,  .741}{\textbf{0.932}} & 　-    & \textcolor[rgb]{1,  0,  0}{\textbf{0.948}} \\
          & MB    & 0.788  & 0.878  & \textbf{0.889} & 　-    & 0.872  & 0.866  & 0.842  & \textcolor[rgb]{.31,  .506,.741}{\textbf{0.912}} & 0.874  & 　-    & 0.869  & 　-    & \textcolor[rgb]{1,  0,  0}{\textbf{0.936}} \\
          & CC    & 0.766  & \textcolor[rgb]{.31,  .506,  .741}{\textbf{0.829}} & 0.764  & 　-    & 0.809  & 0.811  & 0.801  & 0.824  & \textbf{0.826} & 　-    & 0.741  & 　-    & \textcolor[rgb]{1,  0,  0}{\textbf{0.845}} \\
          & JPEG  & 0.735  & 0.757  & 0.800  & 　-    & 0.795  & 0.840  & 0.789  & \textbf{0.847} & 0.763  & 　-    & \textcolor[rgb]{1,  0,  0}{\textbf{0.871}} & 　-    & \textcolor[rgb]{.31,  .506,  .741}{\textbf{0.854}} \\
          & J2K   & 0.843  & 0.815  & 0.789  & 　-    & 0.834  & \textbf{0.849} & \textcolor[rgb]{.31,  .506,  .741}{\textbf{0.861}} & 0.826  & 0.754  & 　-    & 0.834  & 　-    & \textcolor[rgb]{1,  0,  0}{\textbf{0.893}} \\
          & \textbf{Overall} & 0.757  & 0.891  & 0.879  & \textcolor[rgb]{0,  .439,  .753}{\textbf{0.904}} & 0.891  & 0.883  & 0.884  & \textbf{0.899} & 0.852  & 　-    & 0.888  & 　-    & \textcolor[rgb]{1,  0,  0}{\textbf{0.912}} \\
    \midrule
    \multirow{7}[2]{*}{\begin{sideways}SROCC\end{sideways}} & GN    & 0.880  & 0.869  & 0.876  & \textbf{0.893} & 0.891  & 0.880  & 0.888  & 0.867  & 0.872  & 　-    & \textcolor[rgb]{1,  0,  0}{\textbf{0.909}} & 　-    & \textcolor[rgb]{.31,  .506,  .741}{\textbf{0.901}} \\
          & GB    & 0.840  & 0.917  & 0.924  & 0.924  & 0.913  & 0.913  & 0.919  & \textcolor[rgb]{1,  0,  0}{\textbf{0.937}} & 0.863  & 　-    & \textbf{0.927} & 　-    & \textcolor[rgb]{.31,  .506,  .741}{\textbf{0.934}} \\
          & MB    & 0.794  & 0.874  & 0.894  & 0.904  & 0.875  & 0.870  & \textcolor[rgb]{1,  0,  0}{\textbf{0.935}} & \textbf{0.913} & 0.850  & 　-    & 0.872  & 　-    & \textcolor[rgb]{.31,  .506,  .741}{\textbf{0.932}} \\
          & CC    & 0.687  & \textbf{0.722} & 0.611  & 0.665  & 0.713  & 0.704  & 0.664  & \textcolor[rgb]{1,  0,  0}{\textbf{0.762}} & 0.687  & 　-    & 0.526  & 　-    & \textcolor[rgb]{.31,  .506,  .741}{\textbf{0.729}} \\
          & JPEG  & 0.737  & 0.750  & 0.799  & \textcolor[rgb]{.31,  .506,  .741}{\textbf{0.847}} & 0.793  & 0.843  & 0.786  & \textbf{0.846} & 0.718  & 　-    & \textcolor[rgb]{1,  0,  0}{\textbf{0.870}} & 　-    & 0.837  \\
          & J2K   & 0.834  & 0.812  & 0.783  & \textbf{0.862} & 0.828  & 0.844  & \textcolor[rgb]{.31,  .506,  .741}{\textbf{0.862}} & 0.816  & 0.744  & 　-    & 0.860  & 　-    & \textcolor[rgb]{1,  0,  0}{\textbf{0.899}} \\
          & \textbf{Overall} & 0.742  & 0.880  & 0.863  & \textcolor[rgb]{0,  .439,  .753}{\textbf{0.894}} & 0.884  & 0.874  & 0.882  & \textbf{0.890} & 0.832  & 　-    & 0.877 & 　-    & \textcolor[rgb]{1,  0,  0}{\textbf{0.906}} \\
    \midrule
    \multirow{7}[2]{*}{\begin{sideways}RMSE\end{sideways}} & GN    & 6.566  & 6.876  & 6.827  & 　-    & 6.404  & 6.684  & 6.558  & 7.202  & \textbf{6.268} & 　-    & \textcolor[rgb]{.31,  .506,  .741}{\textbf{5.891}} & 　-    & \textcolor[rgb]{1,  0,  0}{\textbf{5.570}} \\
          & GB    & 8.236  & \textbf{5.592} & 5.827  & 　-    & 6.155  & 6.146  & 5.964  & \textcolor[rgb]{.31,  .506,  .741}{\textbf{5.519}} & 6.738  & 　-    & 6.059  & 　-    & \textcolor[rgb]{1,  0,  0}{\textbf{4.562}} \\
          & MB    & 8.006  & 6.236  & \textbf{5.964} & 　-    & 6.360  & 6.518  & 7.012  & \textcolor[rgb]{.31,  .506,  .741}{\textbf{5.335}} & 6.466  & 　-    & 6.327  & 　-    & \textcolor[rgb]{1,  0,  0}{\textbf{4.447}} \\
          & CC    & 8.088  & \textbf{7.048} & 8.114  & 　-    & 7.400  & 7.364  & 7.528  & 7.132  & \textcolor[rgb]{.31,  .506,  .741}{\textbf{6.874}} & 　-    & 8.143  & 　-    & \textcolor[rgb]{1,  0,  0}{\textbf{6.367}} \\
          & JPEG  & 6.372  & 6.143  & 5.640  & 　-    & 5.697  & 5.101  & 5.779  & \textbf{4.994} & 5.862  & 　-    & \textcolor[rgb]{1,  0,  0}{\textbf{4.509}} & 　-    & \textcolor[rgb]{.31,  .506,  .741}{\textbf{4.571}} \\
          & J2K   & 5.589  & 6.023  & 6.388  & 　-    & 5.731  & \textbf{5.499} & \textcolor[rgb]{.31,  .506,  .741}{\textbf{5.293}} & 5.864  & 6.501  & 　-    & 6.127  & 　-    & \textcolor[rgb]{1,  0,  0}{\textbf{4.369}} \\
          & \textbf{Overall} & 9.352  & \textbf{6.499} & 6.831  & \textcolor[rgb]{0,  .439,  .753}{\textbf{6.115}} & 6.503  & 6.723  & 6.695  & \textbf{6.260} & 7.415 & 　-    & 6.778 & 　-    & \textcolor[rgb]{1,  0,  0}{\textbf{5.934}} \\
    \bottomrule
    \bottomrule
    \end{tabular}}%
  \label{tab:crossSIQAD}%
\end{table*}%

\begin{table*}[!t]
  \centering
  \caption{Continued cross-dataset quantitative comparisons under different distortion types on the SCID dataset.}
     \vspace{-0.2cm}
  \resizebox{\linewidth}{!}{
    \begin{tabular}{ccccccccccccccc}
    \toprule
    \toprule
          & \multirow{2}[1]{*}{Distortion} & PieAPP~\cite{Prashnani2018PieAPP} & SQMS~\cite{Gu2016Saliency} & GSS~\cite{Ni2016Gradient} & SFUW~\cite{Fang2017Objective}  & ESIM~\cite{Ni2017ESIM}  & CNN-SQE~\cite{Zhang2018Quality} & SVQI~\cite{Gu2017Evaluating}  & GFM~\cite{Ni2018A}   & EFGD~\cite{2019WangScreen}  & QODCNN-FR~\cite{Jiang2019DeepOM} & RIQA~\cite{jiang2019no}  & SIQA-DF-II~\cite{jiang2019screen} & SR-CNN \\
          &       & 2018  & 2016  & 2016  & 2017  & 2017  & 2018  & 2018  & 2018  & 2019  & 2019  & 2019  & 2019  &  \\
     \midrule
    \multirow{7}[0]{*}{\begin{sideways}PLCC\end{sideways}} & GN    & 0.898  & 0.930  & 0.794  & 　-    & \textbf{0.956} & 　-    & 0.936  & 0.950  & 0.949  & \textcolor[rgb]{1,  0,  0}{\textbf{0.960}} & 　-    & 　-    & \textcolor[rgb]{.31,  .506,  .741}{\textbf{0.960}} \\
          & GB    & 0.874  & 0.908  & 0.887  & 　-    & 0.870  & 　-    & \textbf{0.913} & \textcolor[rgb]{.31,  .506,  .741}{\textbf{0.916}} & 0.909  & 0.866  & 　-    & 　-    & \textcolor[rgb]{1,  0,  0}{\textbf{0.938}} \\
          & MB    & 0.787  & 0.897  & 0.888  & 　-    & 0.882  & 　-    & \textbf{0.900} & \textcolor[rgb]{.31,  .506,  .741}{\textbf{0.902}} & 0.894  & 0.849  & 　-    & 　-    & \textcolor[rgb]{1,  0,  0}{\textbf{0.921}} \\
          & CC    & 0.749  & 0.844  & 0.627  & 　-    & 0.791  & 　-    & 0.827  & \textcolor[rgb]{.31,  .506,  .741}{\textbf{0.879}} & \textbf{0.858} & 0.817  & 　-    & 　-    & \textcolor[rgb]{1,  0,  0}{\textbf{0.890}} \\
          & JPEG  & 0.770  & 0.930  & 0.886  & 　-    & \textbf{0.942} & 　-    & 0.936  & 0.939  & \textcolor[rgb]{.31,  .506,  .741}{\textbf{0.949}} & 0.942  & 　-    & 　-    & \textcolor[rgb]{1,  0,  0}{\textbf{0.958}} \\
          & J2K   & 0.843  & \textbf{0.947} & 0.849  & 　-    & 0.946  & 　-    & \textcolor[rgb]{.31,  .506,  .741}{\textbf{0.951}} & 0.923  & 0.945  & 0.940  & 　-    & 　-    & \textcolor[rgb]{1,  0,  0}{\textbf{0.959}} \\
          & \textbf{Overall} & 0.803  & 0.856  & 0.774  & 0.859  & 0.863  & \textcolor[rgb]{0,  .439,  .753}{\textbf{0.915}} & 0.860  & 0.876  & \textbf{0.885} & 0.882  & 　-    & 0.851 & \textcolor[rgb]{1,  0,  0}{\textbf{0.917}} \\
    \midrule
    \multirow{7}[1]{*}{\begin{sideways}SROCC\end{sideways}} & GN    & 0.880  & 0.916  & 0.697  & 0.929  & \textbf{0.946} & \textcolor[rgb]{.31,  .506,  .741}{\textbf{0.949}} & 0.919  & 0.937  & 0.940  & 0.938  & 　-    & 　-    & \textcolor[rgb]{ 1,  0,  0}{\textbf{0.950}} \\
          & GB    & 0.846  & \textbf{0.908} & 0.887  & 0.897  & 0.870  & 0.907  & \textbf{0.908} & \textcolor[rgb]{.31,  .506,  .741}{\textbf{0.908}} & 0.905  & 0.856  & 　-    & 　-    & \textcolor[rgb]{1,  0,  0}{\textbf{0.938}} \\
          & MB    & 0.783  & 0.881  & 0.873  & 0.812  & 0.861  & 0.878  & \textbf{0.884} & \textcolor[rgb]{.31,  .506,  .741}{\textbf{0.889}} & 0.869  & 0.828  & 　-    & 　-    & \textcolor[rgb]{1,  0,  0}{\textbf{0.898}} \\
          & CC    & 0.735  & \textbf{0.803} & 0.578  & 0.744  & 0.618  & 0.745  & 0.771  & \textcolor[rgb]{.31,  .506,  .741}{\textbf{0.823}} & 0.784  & 0.687  & 　-    & 　-    & \textcolor[rgb]{1,  0,  0}{\textbf{0.824}} \\
          & JPEG  & 0.707  & 0.924  & 0.884  & 0.900  & \textbf{0.946} & \textcolor[rgb]{.31,  .506,  .741}{\textbf{0.947}} & 0.929  & 0.928  & 0.944  & 0.935  & 　-    & 　-    & \textcolor[rgb]{1,  0,  0}{\textbf{0.951}} \\
          & J2K   & 0.835  & 0.932  & 0.836  & 0.910  & 0.936  & \textcolor[rgb]{.31,  .506,  .741}{\textbf{0.939}} & \textbf{0.937} & 0.909  & 0.933  & 0.916  & 　-    & 　-    & \textcolor[rgb]{1,  0,  0}{\textbf{0.942}} \\
          & \textbf{Overall} & 0.787  & 0.832  & 0.775  & 0.856  & 0.848  & \textcolor[rgb]{0,  .439,  .753}{\textbf{0.914}} & 0.839  & 0.876  & \textbf{0.877} & 0.876  & 　-    & 0.851 & \textcolor[rgb]{1,  0,  0}{\textbf{0.915}} \\
    \midrule
    \multirow{7}[2]{*}{\begin{sideways}RMSE\end{sideways}} & GN    & 6.566  & 4.625  & 7.647  & 　-    & \textcolor[rgb]{.31,  .506,  .741}{\textbf{3.676}} & 　-    & 4.418  & \textbf{3.938} & 3.963  & 　-    & 　-    & 　-    & \textcolor[rgb]{1,  0,  0}{\textbf{3.430}} \\
          & GB    & 6.119  & 4.434  & 4.888  & 　-    & 5.221  & 　-    & \textbf{4.319} & \textcolor[rgb]{.31,  .506,  .741}{\textbf{4.257}} & 4.419  & 　-    & 　-    & 　-    & \textcolor[rgb]{1,  0,  0}{\textbf{3.597}} \\
          & MB    & 6.530  & 4.835  & 5.035  & 　-    & 5.143  & 　-    & \textbf{4.771} & \textcolor[rgb]{.31,  .506,  .741}{\textbf{4.612}} & 4.905  & 　-    & 　-    & 　-    & \textcolor[rgb]{1,  0,  0}{\textbf{4.177}} \\
          & CC    & 7.248  & 4.800  & 6.976  & 　-    & 5.479  & 　-    & 5.037  & \textcolor[rgb]{.31,  .506,  .741}{\textbf{4.273}} & \textbf{4.599} & 　-    & 　-    & 　-    & \textcolor[rgb]{1,  0,  0}{\textbf{4.005}} \\
          & JPEG  & 5.710  & 5.518  & 6.984  & 　-    & \textbf{5.037} & 　-    & 5.305  & 5.201  & \textcolor[rgb]{.31,  .506,  .741}{\textbf{4.727}} & 　-    & 　-    & 　-    & \textcolor[rgb]{1,  0,  0}{\textbf{4.195}} \\
          & J2K   & 8.084  & \textbf{5.119} & 8.398  & 　-    & 5.170  & 　-    & \textcolor[rgb]{.31,  .506,  .741}{\textbf{4.906}} & 6.139  & 5.220  & 　-    & 　-    & 　-    & \textcolor[rgb]{1,  0,  0}{\textbf{4.313}} \\
          & \textbf{Overall} & \textcolor[rgb]{.31,  .506,  .741}{\textbf{5.661}} & 7.328  & 8.116  & 　-    & 7.155  & 　-    & 7.218  & 6.831 & {\textbf{6.604}} & 　-    & 　-    & 7.069  & \textcolor[rgb]{1,  0,  0}{\textbf{5.284}} \\
    \bottomrule
    \bottomrule
    \end{tabular}}%
  \label{tab:crossSCID}%
\end{table*}%

\section{Experimental results }
We have evaluated our method over 2 widely adopted datasets, including our previously published SIQAD~\cite{Huan2015Perceptual} dataset, and SCID~\cite{Ni2017ESIM} dataset.

\textbf{SIQAD Dataset}: It contains 20 reference SCIs with diverse visual content and 980 distorted versions generated from the reference SCIs with seven types of image distortions at seven distortion levels, including Gaussian Noise (GN), Gaussian Blur (GB), Motion Blur (MB), Contrast Change (CC), JPEG Compression (JPEG), JPEG2000 Compression (J2K), and the Layer-segmentation based Compression (LSC). The more details about this dataset can be referred to our previous work~\cite{Huan2015Perceptual}.\\
\indent
\textbf{SCID Dataset}: It consists of 40 original SCIs and 1,800 distorted SCIs created with nine different distortion types at five different levels of degradations. The nine distortion types considered are the Gaussian Noise (GN), Gaussian Blur (GB), Motion Blur (MB), Contrast Change (CC), Color Saturation Change (CSC), Color Quantization with Dithering (CQD), JPEG Compression (JPEG), JPEG2000 Compression (J2K), and HEVC Screen Content Compression (HEVC-SCC). The resolution of all SCIs in this dataset is $1280\times720$.\\
\indent
In the SCID dataset, considering no visual distortion is introduced when the images are viewed in grayscale with the distortion of color saturation change. Thus, we exclude CSC distortion in our work.
Also, we have adopted 3 widely adopted evaluation metrics, including PLCC, SROCC and RMSE, and more details can be found in our previous work~\cite{Huan2015Perceptual}.
\subsection{Evaluation Metrics}
The predicted quality scores by different IQA methods might be nonlinearly related to the subjective scores (i.e., MOS: Mean Opinion Score or DMOS: Differential Mean Opinion Score), we employ a five-parameter logistic transform function to map the quality scores to the MOS/DMOS values as follow.\\
\begin{equation}
f(x)=\beta_1\cdot(\frac{1}{2}-\frac{1}{1+e^{(\beta_2\cdot(x-\beta_3))}})+\beta_4\cdot x+\beta_5
\end{equation}
where $\beta_1$, $\beta_2$, $\beta_3$, $\beta_4$ and $\beta_5$ are parameters determined by objective scores and subjective scores, and more details can be found in our previous work~\cite{Huan2015Perceptual}.\\
\indent
After the above-mentioned non-linear regression, 3 widely adopted performance criteria are used to evaluate the proposed SR-CNN model and other quality assessment methods: i.e., Pearson Linear Correlation Coefficient (PLCC), Spearman Rank-order Correlation Coefficient (SROCC), and Root Mean Squared Error (RMSE). The PLCC is defined as:\\
\begin{equation}
{\rm PLCC}=\frac{\sum_{\substack i=1}^N(O_i-\bar{O})(S_i-\bar{S})}{\sqrt{\sum_{\substack i=1}^N(O_i-\bar{O})\cdot\sum_{\substack i=1}^N(S_i-\bar{S})}}
\end{equation}
where $O_i$ and $S_i$ are the objective and subjective scores of the i-th image separately in the database, $\bar{O}$ and $\bar{S}$ are the corresponding mean values of $O_i$ and $S_i$, $N$ denotes the total number of the images in the dataset.\\
\indent
We can calculate the SROCC as following:
\begin{equation}
{\rm SROCC}=1-\frac{6\sum_{\substack i=1}^Nd_i^2}{N(N^2-1)}
\end{equation}
where $d_i$ is the difference between the i-th image's rank in the subjective and objective result, respectively.\\
\indent
The RMSE can be computed as following:
\begin{equation}
{\rm RMSE}=\sqrt{\frac{\sum_{\substack i=1}^N(O_i-S_i)^2}{N}}
\end{equation}
\indent
For a superior correlation between the objective and the subjective scores, PLCC=SROCC=1, and RMSE=0. Thus, the higher the values of the PLCC, SROCC and the lower RMSE value, the better the performance of the quality assessment metric.

\subsection{Quantitative Comparisons}
To demonstrate the performance superiority, we firstly compare the proposed SR-CNN with 4 deep learning based SOTA IQA methods for SCIs, including CNN-SQE18~\cite{Zhang2018Quality}, QODCNN-FR19~\cite{Jiang2019DeepOM}, RIQA19~\cite{jiang2019no} and SIQA-DF-II19~\cite{jiang2019screen}.


Since all these deep learning based methods have adopted the supervised training protocol, we randomly divide the adopted dataset into two non-overlapping subsets by reference images: $80\%$ for training and the rest $20\%$ for testing.
We have repeated such 80\%-20\% partition 50 times, and show the overall performance comparisons in Tab.~\ref{tab:overall}, where the top-3 bests are highlighted in red, blue and black bold, respectively.
Meanwhile, because the codes of these deep learning based SOTA methods are currently not public available, we refer the quantitative results of these methods from the original papers.
In Tab..~\ref{tab:overall}, our proposed SR-CNN method significantly outperforms all these SOTA methods over the SIQAD dataset in terms of all measurements (i.e., PLCC, SROCC and RMSE).
Also, our method is also leading the performance over the SCID dataset (see Tab.~\ref{tab:crossSCID}).

\vspace{-0.2cm}

\subsection{Cross-Dataset Evaluation}
We also use the cross-dataset validations (using the entire SIQAD or SCID as training set) to verify the generalization ability of the proposed learning model, in which we have conducted an extensive cross-dataset comparisons between the proposed method and 12 SOTA methods, including PieAPP~\cite{Prashnani2018PieAPP}, SQMS~\cite{Gu2016Saliency}, GSS~\cite{Ni2016Gradient}, SFUW~\cite{Fang2017Objective}, ESIM~\cite{Ni2017ESIM}, CNN-SQE~\cite{Zhang2018Quality}, SVQI~\cite{Gu2017Evaluating}, GFM~\cite{Ni2018A}, EFGD~\cite{2019WangScreen}, QODCNN-FR~\cite{Jiang2019DeepOM}, RIQA~\cite{jiang2019no} and SIQA-DF-II~\cite{jiang2019screen}.

Considering that the adopted two datasets contain different distortion types, we only show the evaluation results over 6 common distortion types, including GN, GB, MB, CC, JPEG and J2K.
We show the detailed cross-dataset evaluation results in Tab.~\ref{tab:crossSIQAD} and Tab.~\ref{tab:crossSCID}, the top three results are highlighted in red, blue, and black respectively, which indicate that our method outperforms all the compared SOTA methods.

\begin{table*}[!t]
  \centering
   \caption{The cross-dataset result comparison of DL$^+$ and SR-CNN on the SIQAD and SCID}
   \vspace{-0.2cm}
  \resizebox{0.8\linewidth}{!}{
    \begin{tabular}{c|ccccccc|ccccccc}
    \toprule
        \toprule
          & \multicolumn{7}{c|}{\textbf{SIQAD Dataset}}                            & \multicolumn{7}{c}{\textbf{SCID Dataset}} \\
    Distortion & GN    & GB    & MB    & CC    & JPEG  & J2K   & Overall & GN    & GB    & MB    & CC    & JPEG  & J2K   & Overall \\
        \midrule
          & \multicolumn{7}{c|}{PLCC}                             & \multicolumn{7}{c}{PLCC} \\
    DL$^+$   & 0.900  & 0.930  & 0.909  & 0.822  & 0.850  & 0.880  & 0.907  & 0.952  & 0.936  & 0.914  & 0.881  & 0.951  & 0.958  & 0.915  \\
    SR-CNN & 0.922  & 0.948  & 0.936  & 0.845  & 0.854  & 0.893  & 0.912  & 0.960  & 0.938  & 0.921  & 0.890  & 0.958  & 0.959  & 0.917  \\
          & \multicolumn{7}{c|}{SROCC}                            & \multicolumn{7}{c}{SROCC} \\
    DL$^+$   & 0.891  & 0.929  & 0.911  & 0.735  & 0.847  & 0.882  & 0.904  & 0.937  & 0.936  & 0.880  & 0.798  & 0.946  & 0.938  & 0.915  \\
    SR-CNN & 0.901  & 0.934  & 0.932  & 0.729  & 0.837  & 0.899  & 0.906  & 0.950  & 0.938  & 0.898  & 0.824  & 0.951  & 0.942  & 0.915  \\
          & \multicolumn{7}{c|}{RMSE}                             & \multicolumn{7}{c}{RMSE} \\
    DL$^+$   & 6.512  & 5.574  & 5.412  & 7.173  & 4.954  & 4.942  & 6.197  & 3.746  & 3.647  & 4.387  & 4.163  & 4.522  & 4.420  & 5.333  \\
    SR-CNN & 5.570  & 4.562  & 4.447  & 6.367  & 4.571  & 4.369  & 5.934  & 3.430  & 3.597  & 4.177  & 4.005  & 4.195  & 4.313  & 5.284  \\
    \bottomrule
        \bottomrule
    \end{tabular}}%
  \label{tab:gfm}%
\end{table*}%

\vspace{-0.2cm}
\subsection{Parameter and Component Evaluation}
There are many parameters are predefined in our method, including $\alpha_1$, $\alpha_2$, $\beta_1$ and $\beta_2$, which were mentioned in $\rm GSIM_{t/p}^+$ (Eq.~\ref{eq:GSIMCW}) and $\rm FSIM$ (Eq.~\ref{eq:ImageScore}), respectively.
So, we resort a extensive ablation study on these parameters to facilitate an optimal choice.
We initially fixed $\beta=1$ and simply tested multiple choices of $\alpha_1$ (ranging from 0.1 to 0.5) and $\alpha_2$ (ranging from 0 to 0.8), and assign $\alpha_1=0.2$ and $\alpha_2=0.6$ as the optimal choice according to Tab.~\ref{tab:parameters1} and Tab.~\ref{tab:parameters2}.
\begin{table}[htbp]
   \centering
   \caption{Analysis of GSIM$_{t/p}$$^+$ result with different $\alpha_1$ values on the SIQAD and the SCID}
      \vspace{-0.2cm}
  \resizebox{0.5\linewidth}{!}{
    \begin{tabular}{cccc|ccc}
    \toprule
        \toprule
    \multirow{2}[2]{*}{$\beta_1$=1} & \multicolumn{3}{c|}{SIQAD} & \multicolumn{3}{c}{SCID} \\
          & PLCC  & SROCC & RMSE  & PLCC  & SROCC & RMSE \\
    \midrule
    $\alpha_1$=0.1 & 0.839  & 0.827  & 7.781  & 0.845  & 0.844  & 7.630  \\
    $\alpha_1$=0.2 & \textcolor[rgb]{ 1,  0,  0}{\textbf{  0.843 }} & \textcolor[rgb]{ 1,  0,  0}{\textbf{  0.827 }} & \textcolor[rgb]{ 1,  0,  0}{\textbf{  7.707 }} & \textcolor[rgb]{ 1,  0,  0}{\textbf{  0.846 }} & \textcolor[rgb]{ 1,  0,  0}{\textbf{  0.843 }} & \textcolor[rgb]{ 1,  0,  0}{\textbf{ 7.593 }} \\
    $\alpha_1$=0.3 & 0.840  & 0.820  & 7.762  & 0.814  & 0.811  & 8.285  \\
    $\alpha_1$=0.4 & 0.826  & 0.796  & 8.076  & 0.749  & 0.737  & 9.452  \\
    $\alpha_1$=0.5 & 0.790  & 0.738  & 8.773  & 0.639  & 0.621  & 10.970  \\
    \bottomrule
            \bottomrule
    \end{tabular}}%
  \label{tab:parameters1}%
\end{table}%
\begin{table}[htbp]
  \centering
 \caption{Analysis of FSIM result with different $\alpha_2$ values on the SIQAD and the SCID}
    \vspace{-0.2cm}
  \resizebox{0.5\linewidth}{!}{
    \begin{tabular}{cccc|ccc}
    \toprule
        \toprule
    \multirow{2}[2]{*}{$\beta_2$=1} & \multicolumn{3}{c|}{SIQAD} & \multicolumn{3}{c}{SCID} \\
          & PLCC  & SROCC & RMSE  & PLCC  & SROCC & RMSE \\
    \midrule
    $\alpha_2$=0.0 & 0.843  & 0.827  & 7.707  & 0.846  & 0.843  & 7.593  \\
    $\alpha_2$=0.2 & 0.858  & 0.848  & 7.352  & 0.855  & 0.851  & 7.392  \\
    $\alpha_2$=0.4 & 0.865  & 0.859  & 7.188  & 0.857  & 0.852  & 7.351  \\
    $\alpha_2$=0.6 & \textcolor[rgb]{ 1,  0,  0}{\textbf{  0.867 }} & \textcolor[rgb]{ 1,  0,  0}{\textbf{  0.863 }} & \textcolor[rgb]{ 1,  0,  0}{\textbf{  7.130 }} & \textcolor[rgb]{ 1,  0,  0}{\textbf{  0.857 }} & \textcolor[rgb]{ 1,  0,  0}{\textbf{  0.851 }} & \textcolor[rgb]{ 1,  0,  0}{\textbf{  7.345 }} \\
    $\alpha_2$=0.8 & 0.867  & 0.864  & 7.146  & 0.853  & 0.849  & 7.432  \\
    \bottomrule
        \bottomrule
    \end{tabular}}%
  \label{tab:parameters2}%
\end{table}%

We further conduct component evaluation to validate the effectiveness of each component adopted in our method.
As shown in Tab.~\ref{tab:Components}, the solely local structure similarity (Eq.~\ref{eq:LSIM}) based measurement (denoted by LSIM) exhibits the worst performance, which is even slightly worse than the solely global structure similarity based measurement (denoted by GSIM).
Then, the overall performance can get a significant improvement by dividing the input SCI into texture and picture regions beforehand, which we denote it as $\rm GSIM_{t/p}$.
Further, the overall performance can get further enhanced by measuring the structure similarity from the dictionary perspective (Eq.~\ref{eq:GSIMCW}), and we denote its performance as $\rm GSIM_{t/p}^+$.
By using the multiplicative based fusion to integrate the $\rm GSIM_{t/p}^+$ into the local structure similarity measurement (Eq.~\ref{eq:PatchScore}), the overall performance get a further improvement, which we denote it as FSIM (Eq.~\ref{eq:ImageScore}).
Also, our newly designed selective fusion network is capable of significantly improving the overall performance ($\rm DL$), which simply feeds the ``resized'' feature maps into CNN based deep network.
Specially, our mid-level patch strategy (Sec.~\ref{sec:PAS}) is able to avoid the ``resize'' operation induced additional distortions effectively, and we denote its performance by $\rm DL^{+}$.
Meanwhile, the integrated off-the-shelf texture similarity (Eq.~\ref{eq:UpdatedFinalScore}) also benefits our method slightly, and we denote its performance as our complete version, i.e., the SR-CNN, achieving the best performance.
\begin{table}[htbp]
  \centering
  \caption{The detailed component quantitative evaluation results}
     \vspace{-0.2cm}
  \resizebox{0.6\linewidth}{!}{
    \begin{tabular}{l|ccc|ccc}
    \toprule
        \toprule
    \multirow{2}[2]{*}{Components} & \multicolumn{3}{c|}{\textbf{SIQAD Dataset}} & \multicolumn{3}{c}{\textbf{SCID Dataset}} \\
          & PLCC  & SROCC & RMSE  & PLCC  & SROCC & RMSE \\
    \midrule
    LSIM  & 0.754  & 0.759  & 9.400  & 0.772  & 0.750  & 9.062  \\
    GSIM  & 0.758  & 0.734  & 9.329  & 0.779  & 0.757  & 8.984  \\
    GSIM$_{t/p}$ & 0.834  & 0.821  & 7.908  & 0.842  & 0.838  & 7.681  \\
    GSIM$_{t/p}$$^+$ & 0.843  & 0.827  & 7.707  & 0.846  & 0.843  & 7.593  \\
    FSIM  & 0.867  & 0.863  & 7.130  & 0.877  & 0.872  & 7.038  \\
    DL    & 0.821  & 0.806  & 8.516  & 0.832  & 0.818  & 8.254  \\
    DL$^+$   & 0.912  & 0.904  & 5.792  & 0.936  & 0.937  & 4.964  \\
    SR\_CNN & \textcolor[rgb]{ 1,  0,  0}{\textbf{ 0.916 }} & \textcolor[rgb]{ 1,  0,  0}{\textbf{ 0.908 }} & \textcolor[rgb]{ 1,  0,  0}{\textbf{ 5.683 }} & \textcolor[rgb]{ 1,  0,  0}{\textbf{ 0.939 }} & \textcolor[rgb]{ 1,  0,  0}{\textbf{ 0.940 }} & \textcolor[rgb]{ 1,  0,  0}{\textbf{ 4.830 }} \\
    \bottomrule
        \bottomrule
    \end{tabular}}%
  \label{tab:Components}%
\end{table}%

Moreover, we have also tested the performance of the CNN model which using the RGB information only (without using both LSIM and GSIM features).
As shown in Tab.~\ref{tab:RGB}, our newly designed hand-crafted features can effectively boost the solely RGB information trained RGB-CNN model significantly, e.g., almost 6\% overall PLCC and SROCC over the SCID dataset.

In addition, we also make cross-dataset experiments to compare performances of the $\rm DL^{+}$ and SR-CNN mentioned in Eq.~\ref{eq:UpdatedFinalScore}. The detailed cross-dataset validation results are shown in Tab.~\ref{tab:gfm}, which further verified the effectiveness of our integrated off-the-shelf texture similarity.
\begin{table*}[!t]
  \centering
  \caption{The quantitative evaluation between the RGB trained CNN model and the proposed SR-CNN on the SIQAD and SCID datasets.}
     \vspace{-0.2cm}
   \resizebox{\linewidth}{!}{
    \begin{tabular}{c|cccccccc|ccccccccc}
    \toprule
        \toprule
          & \multicolumn{8}{c|}{\textbf{SIQAD Dataset}}                                    & \multicolumn{9}{c}{\textbf{SCID Dataset}} \\
    Distortion & GN    & GB    & MB    & CC    & JPEG  & J2K   & LSC   & Overall & GN    & GB    & MB    & CC    & JPEG  & J2K   & HEV. & CQD   & Overall \\
    \midrule
          & \multicolumn{8}{c|}{PLCC}                                     & \multicolumn{9}{c}{PLCC} \\
    RGB-CNN & 0.914  & 0.933  & 0.885  & 0.821  & 0.841  & 0.766  & 0.804  & 0.898  & 0.959  & 0.882  & 0.869  & 0.824  & 0.912  & 0.874  & 0.838  & 0.827  & 0.872  \\
    SR-CNN & 0.916  & 0.935  & 0.912  & 0.852  & 0.866  & 0.872  & 0.875  & 0.916  & 0.967  & 0.938  & 0.938  & 0.877  & 0.963  & 0.961  & 0.918  & 0.922  & 0.939  \\
          & \multicolumn{8}{c|}{SROCC}                                    & \multicolumn{9}{c}{SROCC} \\
    RGB-CNN & 0.891  & 0.918  & 0.883  & 0.718  & 0.828  & 0.755  & 0.792  & 0.885  & 0.950  & 0.883  & 0.868  & 0.656  & 0.897  & 0.844  & 0.750  & 0.781  & 0.869  \\
    SR-CNN & 0.896  & 0.928  & 0.910  & 0.748  & 0.860  & 0.874  & 0.875  & 0.908  & 0.957  & 0.935  & 0.924  & 0.810  & 0.957  & 0.940  & 0.885  & 0.896  & 0.940  \\
          & \multicolumn{8}{c|}{RMSE}                                     & \multicolumn{9}{c}{RMSE} \\
    RGB-CNN & 5.852  & 5.150  & 5.664  & 6.718  & 4.838  & 6.107  & 4.886  & 6.164  & 3.460  & 4.876  & 5.241  & 4.958  & 6.105  & 7.517  & 7.585  & 7.122  & 6.928  \\
    SR-CNN & 5.733  & 5.089  & 5.131  & 6.205  & 4.496  & 4.856  & 3.915  & 5.683  & 3.127  & 3.558  & 3.679  & 4.152  & 3.966  & 4.225  & 5.249  & 4.856  & 4.830  \\
    \bottomrule
        \bottomrule
    \end{tabular}}%
  \label{tab:RGB}%
\end{table*}%

Meanwhile, we have conducted the ablation study to select an optimal patch number ($n$), mentioned in our patch adaptively selection strategy (Sec.~\ref{sec:PAS}).
The ablation results can be found in Tab.~\ref{tab:patches}, which indicate that the parameter $n$ is positively related to the overinsurance.
Since the performance improvement tendency is shrinking when $n\geq 6$, we assign $n=6$ as the optimal choice to strike the trade-off between performance and computation.
\begin{table}[!t]
  \centering
  \caption{Mid-level patch selection ablation study.}
     \vspace{-0.2cm}
  \resizebox{0.5\linewidth}{!}{
    \begin{tabular}{c|cccc}
    \toprule
        \toprule
    number & 0 patches & 3 patches & 6 patches & 9 patches \\
    \midrule
    PLCC  & 0.821  & 0.914  & 0.916  & 0.914  \\
    SROCC & 0.806  & 0.906  & 0.908  & 0.907  \\
    RMSE  & 8.516  & 5.687  & 5.683  & 5.692  \\
    \bottomrule
        \bottomrule
    \end{tabular}}%
  \label{tab:patches}%
\end{table}%

\subsection{Limitation}
\indent	
In addition to the accuracy as discussed above, the computational complexity of the IQA model is another important aspect to be assessed.
Here we list the single image computation time comparisons over the SIQAD dataset in Tab.~\ref{tab:runtime}.
All of these tests are performed by running Matlab R2016a on a desktop computer with i7-6700 3.40GHz CPU, GTX 1080 GPU, 32GB RAM (all the hand-crafted pre-processing stages are conducted on CPU, yet the selective deep fusion is performed on GPU).
Actually, our proposed SR-CNN is relatively time-consuming, which is mainly induced by the heavy time computation cost of the adopted OMP re-construction procedure (almost 17s per-image).
\begin{table}[htbp]
  \centering
  \caption{Runtime comparison (seconds) between 6 most representative methods on the SIQAD dataset.}
     \vspace{-0.2cm}
\resizebox{0.6\linewidth}{!}{
     \begin{tabular}{ccccccc}
    \toprule
        \toprule
    Model  & SFUW17  & ESIM17 & CNN-SQE18 & SVQI18  & GFM18  & SR-CNN \\
    \midrule
    Time/s  & 189.680  & 3.013  & 37.040  & 2.432  & 0.167  & 26.608  \\
    \bottomrule
        \bottomrule
    \end{tabular}}
  \label{tab:runtime}%
\end{table}%

\section{Conclusion}
In this paper, we have proposed a full reference quality assessment
method for SCIs.  The proposed method includes two main stages: 1)
The structure-aware SCI quality assessment stage; 2) The
selective deep fusion stage. At the first stage, we propose to use the
sparse representation to aid a novel global structure similarity
(GSIM) measurement. The conventional local metric is also adopted to
improve its robustness. Then, at the second stage, we adopt the
newly-designed selective deep fusion network to fuse multi-level
structure similarity measurement, achieving an optimal complementary
status. We have conducted extensive quantitative evaluations to
demonstrate the efficacy and effectiveness of our method. Our method
is simple yet effective with a clear motivation, continuing to improve
the current SOTA methods and techniques with new features
and advantages.

\vspace{0.2cm}
\begin{acks}
This research is supported in part by
National Natural Science Foundation of China (No. 61802215
and No. 61806106), Natural Science Foundation of Shandong
Province (No. ZR2019BF011 and ZR2019QF009) and National Science Foundation of USA (No. IIS-1715985 and IIS-
1812606).
\end{acks}

\bibliographystyle{ACM-Reference-Format}
\bibliography{ref1}
\end{document}